# Contact state analysis using NFIS &SOM

H. Owladeghaffari [1]

[1] *Faculty of mining and metallurgical engineering, Amirkabir University of technology, Tehran, Iran*

Email: h.o.ghaffari@gmail.com

**Abstract**  This paper reports application of neuro- fuzzy inference system (NFIS) and self organizing feature map neural networks (SOM) on detection of contact state in a block system. In this manner, on a simple system, the evolution of contact states, by parallelization of DDA, has been investigated. So, a comparison between NFIS and SOM results has been presented. The results show applicability of the proposed methods, by different accuracy, on detection of contact's distribution.

**Key words:**  contact detection, soft computing, DDA

## INTRODUCTION

The aim of contact detection is to report interference between two or more geometric objects in static or dynamic environments.

 The interference report consists of different answers; in some cases: yes or no; in many other cases (such block system) an exact or approximate overlapping extent report is required. The reducing of time consuming in computation of contact detection  in discontinuous deformation analysis(DDA), is the preliminary  aim of this paper.

In this manner, two intelligent Systems: Neuro-Fuzzy Inference System (NFIS) and self-organizing feature map (SOM), has been employed.

Application of fuzzy inference systms  on the static  block system has been highlighted in [1].

Parallelization of contct detaction in DDA, has been recognized in two stages : using NFIS on the obtained data sets from DDA, state of contact between blocks have been evaluated. Then, this procedure has been rendered  on the aggregated data set in the "given window", located in the problem domain, using SOM and fusion of NFIS and SOM.

## FUZZY INFERENCE SYSYTEM

The membership may be described either in a discrete form as a set of membership values or as a continuous function valid over some range of values of the variable x. To the most popular types of membership functions belong the triangle, trapezoidal, Gaussian or bell functions. We have used here the generalized description of the Gaussian function, given in the form $f(x,\sigma,c) = e^{\frac{-(x-c)^2}{2\sigma^2}}$

This is the Gaussian function that depends on two parameters σ and c. The parameters for the Gaussian function represents the parameters σ and c listed in order to the vector.

The most popular solution of the fuzzy networks is based on the so-called fuzzy inference system, fuzzy if-then rules and fuzzy reasoning.

 Such a fuzzy inference system implements a nonlinear mapping from input space to output space. This mapping is accomplished by a number of fuzzy if-then rules, each of which describes the local behaviors of the mapping, like it is done in radial basis function networks.

 The antecedent of the rule defines the fuzzy region in the input space, while the consequent specifies the output of the fuzzy region.

There are different solutions of fuzzy inference systems. Two well known fuzzy modeling methods are the Tsukamoto fuzzy model and Takagi–Sugeno–Kang (TSK) model. In the present work, only the TSK model has been considered. A typical fuzzy rule in this model has the form

If $x_1$ is $A_1$ and $x_2$ is $A_2$ .... and $x_n$ is $A_n$ then $y = f(x)$

Crisp function in the consequent. The function $y = f(x)$ is a polynomial in the input variables $x_1, x_2,..., x_n$. We will apply here the linear form of this function. The aggregated values of the membership function for the vector x may be assumed either in the form of MIN operator or in a product form. For M fuzzy rules of the equation (2), we have M such membership functions $\mu_1, \mu_2,..., \mu_M$. We assume that each antecedent is followed by the consequent of the linear form

$$= p_{i0} + \sum_{j=1}^{n} p_{ij} x_j \quad i = 1, 2... M \text{ and } j = 1, 2... n.$$

The adjusted parameters of the system are the nonlinear parameters ($c_j^{(k)}, \sigma_j^{(k)}, b_j^{(k)}$) for j = 1, 2,..., n and k = 1,2,...,M of the fuzzifier functions and the linear parameters (weights $p_{kj}$) of TSK functions. In contrary to the Mamdani fuzzy inference system, the TSK model generates a crisp output value instead of a fuzzy one. The defuzzifier is not necessary.

$$y(x) = \frac{1}{\sum_{r=1}^{M}\left[\prod_{j=1}^{n}\mu_r(x_j)\right]} \times \sum_{k=1}^{M}\left(\left[\prod_{J=1}^{n}\mu_k(x_j)\right]\left(p_{k0} + \sum_{j=1}^{n}p_{kj}x_j\right)\right)$$

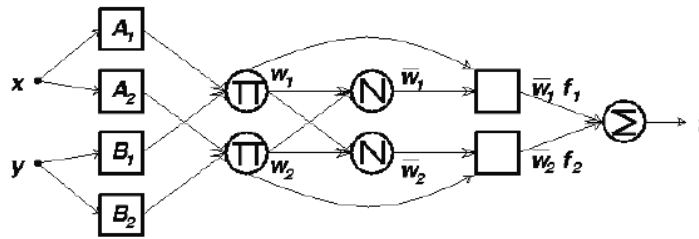

Figure1. A typical ANFIS (TSK) with two inputs and two MF for any input [2].

The TSK fuzzy inference systems can be easily implanted in the form of a so called Neuro-fuzzy network structure. Figure 6 presents the 5-layer structure of a Neuro-fuzzy network, realizing the TSK model of the fuzzy system. It is assumed that the functions $y_i$, $y_i = f_i(x)$ are linear of the form

$$f_i(x) = p_{i0} + \sum_{j=1}^{n} p_{ij} x_j$$

The adaptable parameters of the networks are the variables of the membership functions ($c_j^{(k)}, \sigma_j^{(k)}, b_j^{(k)}$) for j =1,2,...,n, k =1,2,...,M and the coefficients (linear weights) $p_{ij}$ for i =1,2,...,M and j =0,1,2,...,n of the linear Takagi–Sugeno functions. The network in figure 2 has a multilayer form. The first layer performs the fuzzification according to the membership function $\mu_k(x_j)$, described by equation (1). The second layer aggregates the fuzzified results of the individual scalar functions of every variable and determines the membership function of the whole vector x. This is the product type aggregation.

Each node of this layer represents the firing strength of a rule. The third layer calculates the aggregated signal of the fuzzy inference for each inference rule.

The output signal of each unit of this layer is the product of the firing strength of the rule and the consequent membership value.

The fourth layer determines the output membership function. Layer five calculates only the sum of the signal of the third and the second layers of the network.

The final sixth layer contains only one neuron for output. In the case of multiple outputs, we can add as many output neurons as needed in a fashion similar to the case of one output. The output neuron computes the overall output signal according to the equation (4).

One of the most important stages of the Neurofuzzy TSK network generation is the establishment of the inference rules. Often used is the so-called grid method, in which the rules are defined as the combinations of the membership functions for each input variable. If we split the input variable range into a limited number (say $n_i$ for i =1, 2,..., n) of membership functions, the combinations of them lead to many different inference rules.

For example for 10 input systems, at 3 membership functions each, the maximum possible number of rules is equal M = 310 = 59049. The problem is that these combinations correspond in many cases to the regions of no data, and hence a lot of them may be deleted.

This problem can be solved by using the fuzzy self-organization algorithm.

This algorithm splits the data space into a specified number of overlapping clusters.

Each cluster may be associated with the specific rule of the center corresponding to the center of the appropriate cluster.

In this way all rules correspond to the regions of the space-containing majority of data and the problem of the empty rules can be avoided. The ultimate goal of data clustering is to partition the data into similar subgroups. This is accomplished by employing some similar measures (e.g., the Euclidean distance).

In this paper data clustering is used to derive membership functions from measured data, which, in turn, determine the number of If-Then rules in the model (i.e., rules indication).

Several clustering methods have been proposed in the literature [3]. The method employed in this paper is the subtractive clustering method.

The extracted rules from this method can be employed in the cellular automata procedure, to parrallization of block's system computations.

The C.S in 2-D on block system has a four components:" no contact: 0; V-V: 1; V-E: 2; E-E: 3", where numbers are the attributed codes. Figure 2 shows the proposed algorithm on a two block in a static analysis. All of training and checking data set were 100 and 50, respectively, which were revealed from DDA [4]. Inputs for any block were vertexes positions and area (total inputs: 18).

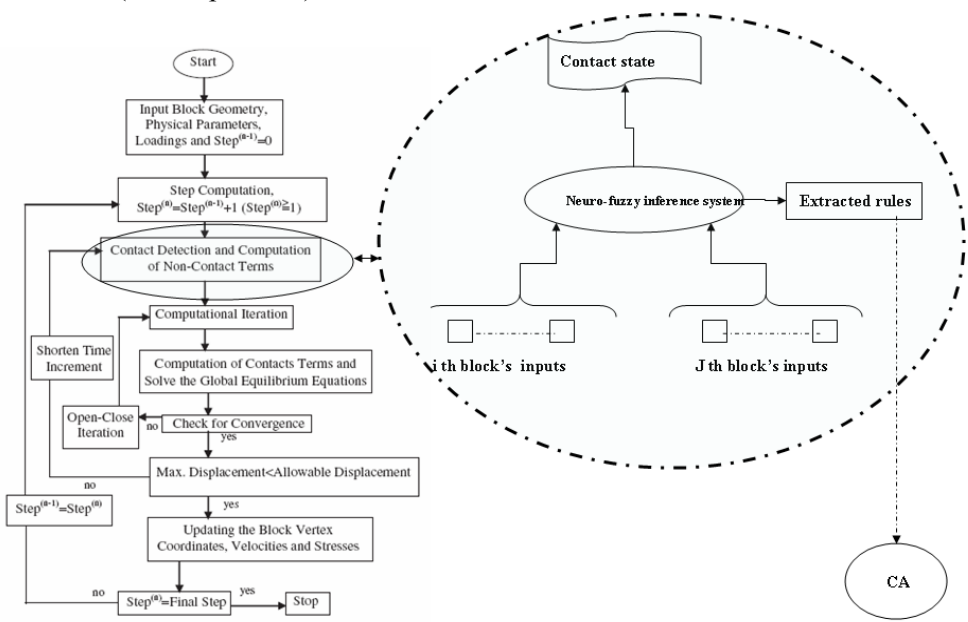

Figure 2. A proposed algorithm based on NFIS &DDA

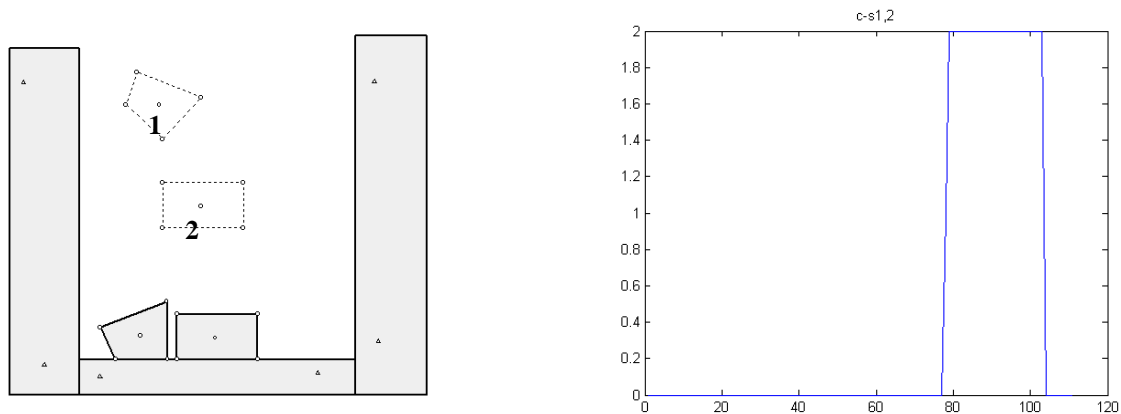

Figure3. the evolution of "contact state" v.s time (in a static analysis on DDA-2D)

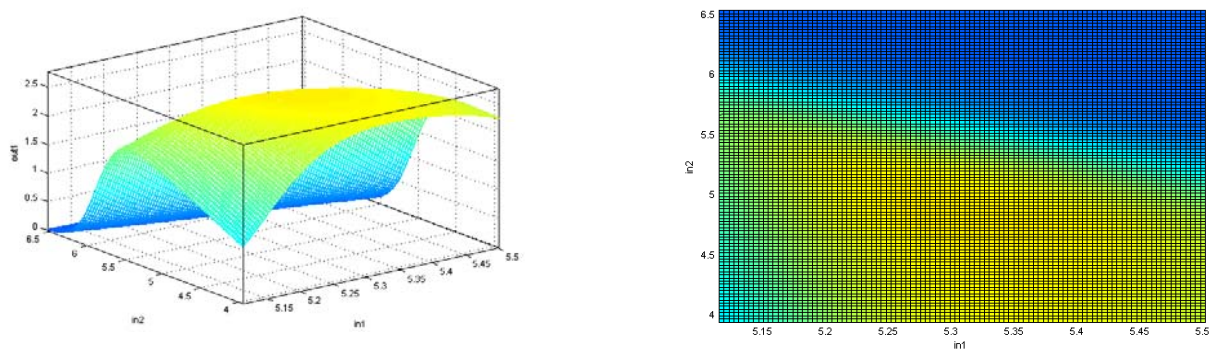

Figure 4(a) .3-D relation between inputs 1, 2 and C.S (block1) using 13 MFs (Gaussian).

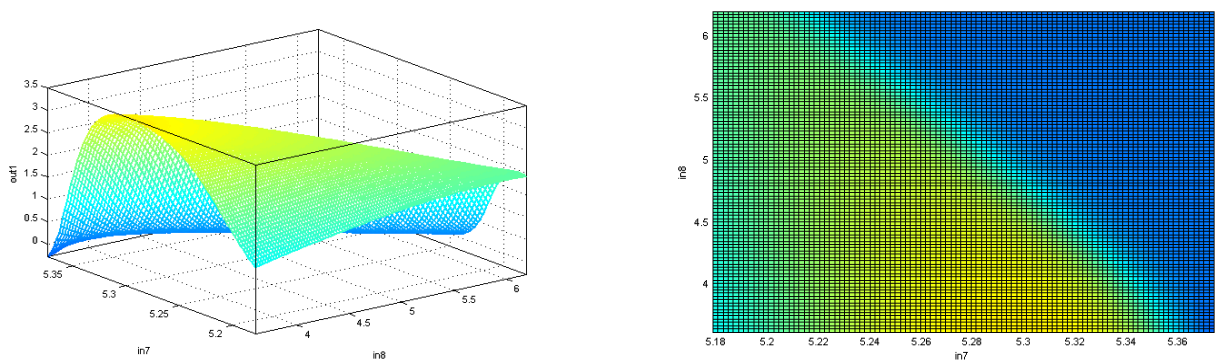

Figure 4(b). 3-D relation between inputs 7, 8 and C.S (block1) using 13 MFs (Gaussian). The cold colors show low C.S while warm colors depicts high C.S in the pseudo colors figures.

The results have been presented in figures 4and 5, which are produced from the selection of 13 and 39 MFs (rules) for any inputs. So, coinciding of the histogram of the same inputs and the extracted MFs ha been shown in figure 6.

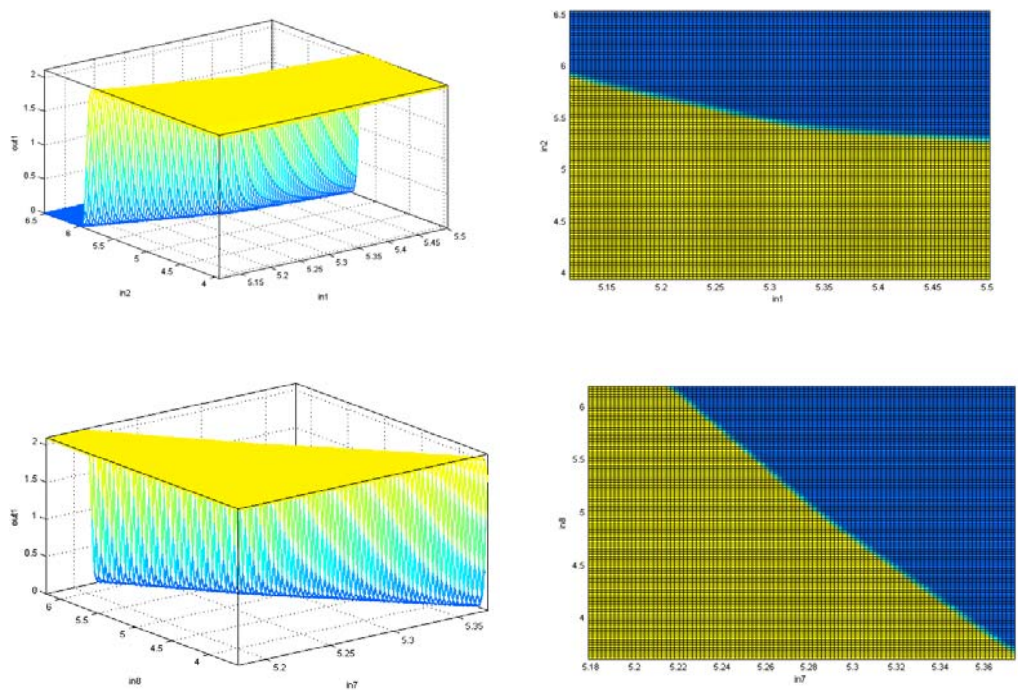

Figure 5 .the extracted relation using 39 MFs; inputs are same as previous

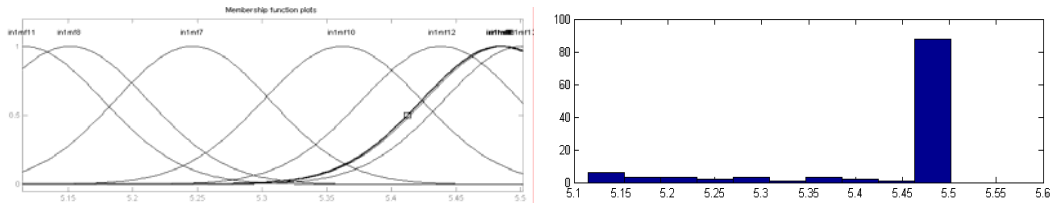

6(a) MFs for input1; using 13 rules for NFIS- histogram distribution of input 1

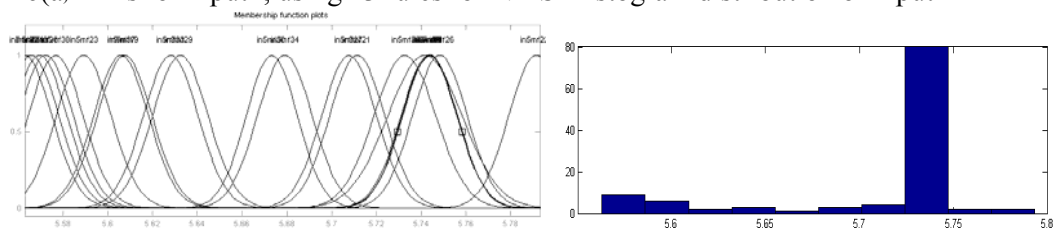

6(b ) MFs for input 5; using 39 rules for NFIS- histogram distribution of input

It is worth noting that decreeing the dominant rules, the accuracy of NFIS responses is diminished

**Self organizing feature map (SOM)**
The modes of information granulation (IG) in which the granules are crisp or fuzzy play important roles in a wide verity of methods, approaches and techniques. One of among them is cluster analysis.[6]

Kohonen self-organizing networks (Kohonen feature maps or topology-preserving maps) are competition-based network paradigm for data clustering. The learning procedure of Kohonen feature maps is similar to the competitive learning networks. The main idea behind competitive learning is simple; the winner takes all. The competitive transfer function returns neural outputs of 0 for all neurons except for the winner which receives the highest net input with output 1.

SOM changes all weight vectors of neurons in the near vicinity of the winner neuron towards the input vector. Due to this property SOM, are used to reduce the dimensionality of complex data (data clustering). Competitive layers will automatically learn to classify input vectors, the classes that the competitive layer finds are depend only on the distances between input vectors([2] , [5]).

Given some "windows" on the block domain, so that their positions were selected,randomly. By scanning of contact state ,within such windows, and extension of procedure to other parts of the domain,using fusion of SOM and NFIS ,can be supposed as an approximated alternative to the previous method (figure 8)

To assessing of 3-D SOM, on the contact state distribution, contact state between 1, 2 (blocks, figure 3) vs. the assembled X& Y directions of the gravity center of any ones, has been considered.

Figure( and table) 7 shows, the performances of 3*3 SOM ( $n_x = 3, n_y = 3$; Matrix of neurons - $n_x$ . $n_y$ determines the size of SOM.) on the assembled X and Y vs. contact state's evolution. So, the coordinates of winner neurons(table) has been depicted. The results have been produced after 300 training epochs.(figure 9)

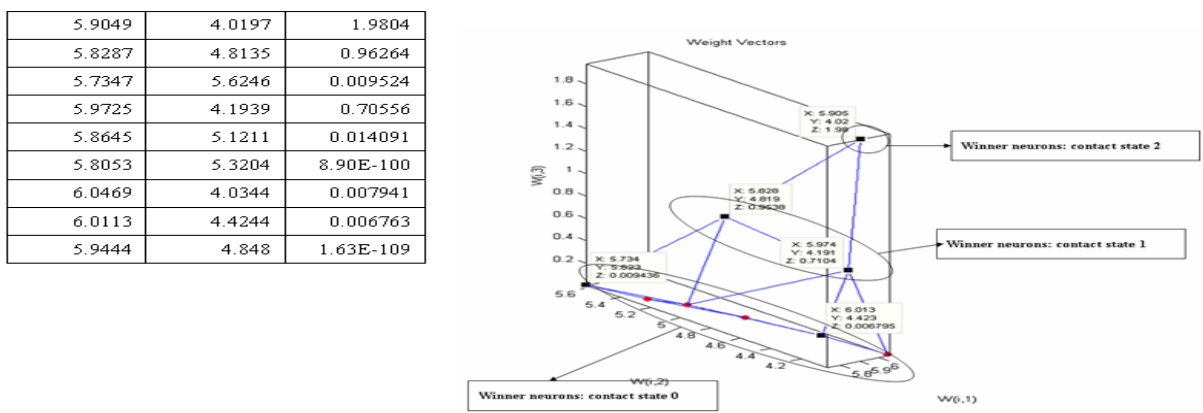

Figure7. Performances of 3-D SOM (nx=3 , ny= 3) on the assembled X, Y vs. C.S(after 300 epoches): this net could recognize all of C.S

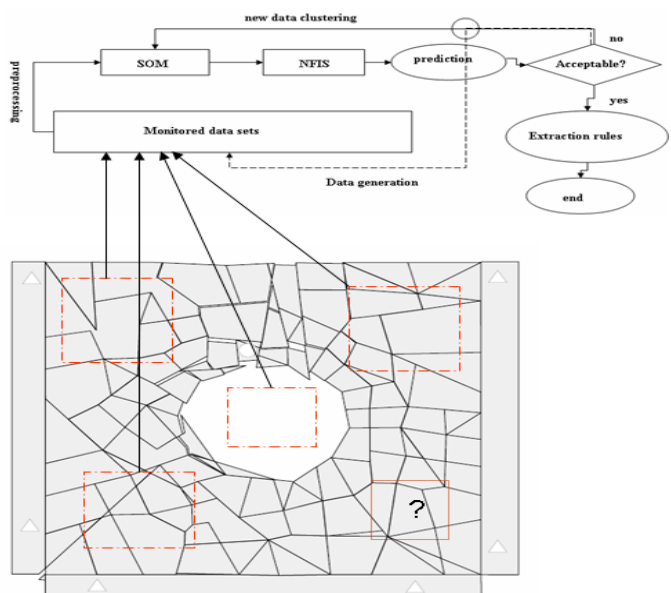

Figure8. genral procedure based on fusion of SOM/NFIS

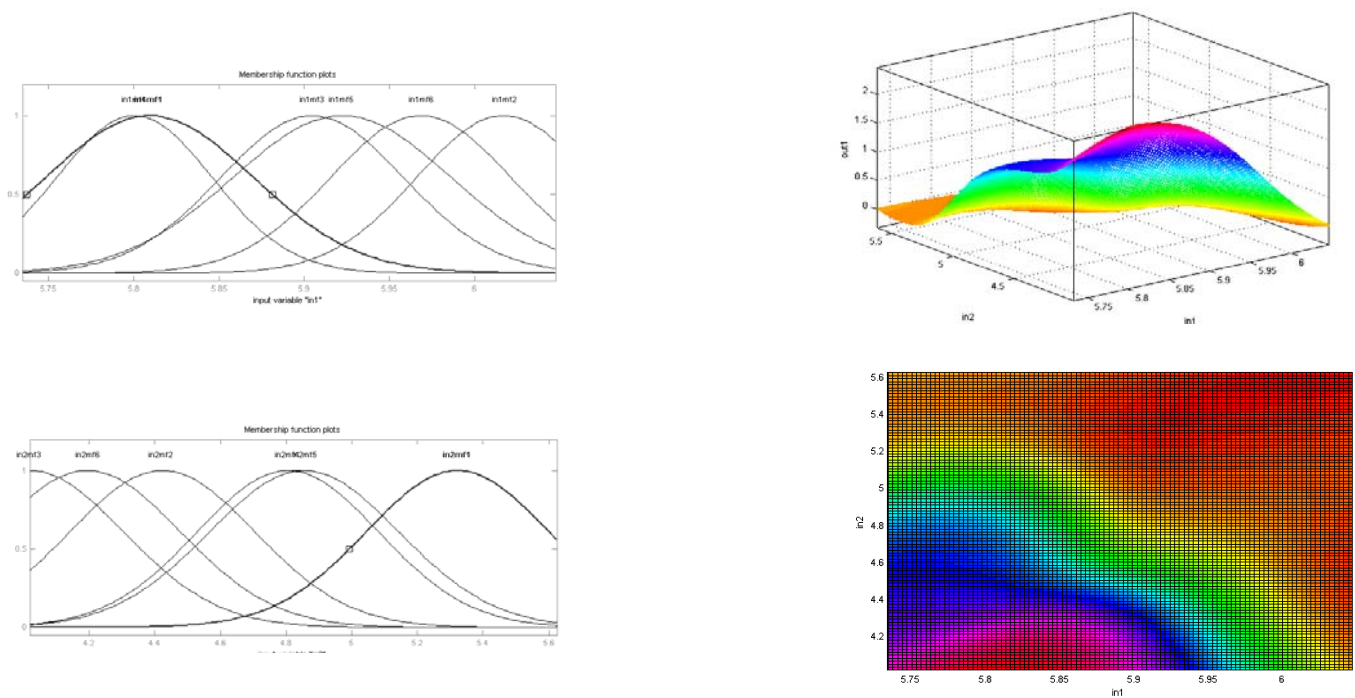

Figure9. performance of SOM/NFIS on the blocks(figure3)

## Conclusion and future work

In this paper, by utilizing two main approaches of computational intelligent, namely, fuzzy set theory and neural networks, some analysis on the contacts of blocks, has been accomplished.Neruo-fuzzy inference system and self organizing feature map neural network, to find contact state, were rendered.

The extraction of rules by NFIS and an acceptable approximation of the distribution of contact's state (resulted from SOM) were two main results of the employed method

Application of fuzzy rules in a cellular automata, clustering of blocks in 2-D as a fuzzy or crisp information granulation and inserting them in to DEM's flowcharts, are the future works.

## Acknowledgements

author would like to thank from Mr.N.Owladeghaffari for his encouragement along this study, so from Miss. M.Seyyedhashemi to assurances, gratefully thanks are recorded.